\pdfoutput=1
\documentclass[letterpaper]{article} 
\usepackage{aaai2027}  
\usepackage[hyphens]{url}  
\usepackage{graphicx} 
\usepackage{natbib}  
\usepackage{caption} 
\usepackage{booktabs}

\usepackage[utf8]{inputenc}
\usepackage{amsfonts}
\usepackage{amsmath}
\usepackage{amssymb}
\usepackage{xcolor}

\definecolor{tablehl}{RGB}{232,244,255}

\newcommand{\ours}{\textsc{GenTrack}}
\newcommand{\tgen}{G_\theta}
\newcommand{\tracker}{\pi_\phi}
\newcommand{\dataset}{\mathcal{D}}

\newcommand{\robotmotion}{\mathbf{q}}
\newcommand{\textprompt}{\mathbf{c}}
\newcommand{\metrichead}[1]{\begin{tabular}[c]{@{}c@{}}#1\end{tabular}}

\title{\ours: Physical Alignment for Robot-Native Motion Generation and Zero-Shot Humanoid Tracking}

\author{
Zeyu Ling\textsuperscript{\rm 1},
Xinyao Yu\textsuperscript{\rm 1},
Renye Yan\textsuperscript{\rm 2},
Jikang Cheng\textsuperscript{\rm 2},\\
Zhanke Wang\textsuperscript{\rm 2},
Qing Shuai\textsuperscript{\rm 3},
Changqing Zou\textsuperscript{\rm 1,\rm 4,*}
}
\affiliations{
\textsuperscript{\rm 1}Zhejiang University,\\
\textsuperscript{\rm 2}Peking University,\\
\textsuperscript{\rm 3}Tencent,\\
\textsuperscript{\rm 4}Zhejiang Lab,\\
\textsuperscript{*}Corresponding author.
}

\begin{document}

\maketitle

\begin{abstract}
General-purpose humanoid trackers can execute diverse references, but their
zero-shot coverage depends on large embodied corpora that are costly to extend.
Text-to-motion generators offer scalable supervision, yet models trained on
human motion or retargeted data inherit a gap between kinematic plausibility
and robot executability.  Existing one-way pipelines fix either the generated
corpus or the reward tracker.  We introduce \ours{}, an online
generator--tracker framework that alternates execution-grounded,
group-relative generator alignment with tracker training on newly generated
references; anchoring and rehearsal constrain drift. On Unitree G1, we
evaluate \ours{} with ProtoMotions and SONIC backbones across three zero-shot
tracking splits including public AMASS and LAFAN benchmarks, and a private
out-of-distribution test set of 1,024 prompt-motion pairs in the wild.
The online co-training strategy consistently produces generators that output
more robot-executable motions with strong semantic alignment, and trackers
with markedly broader zero-shot coverage and improved tracking accuracy,
especially on out-of-distribution references.  These results demonstrate that
joint online post-training effectively narrows the executability gap between
retargeted references and robot-native motion, advancing zero-shot humanoid
control without additional data collection and beyond the limitations of a
static reference pool.
\end{abstract}

\section{Introduction}

A general-purpose humanoid controller should execute reference motions beyond its training set without motion-specific optimization or test-time adaptation.
Recent generalist trackers have advanced zero-shot tracking on diverse motion collections~\citep{yin2025unitracker,han2025kungfubot2,li2026bfmzero,luo2025sonic,ma2026robust,wang2026fast,li2026omnitrack,tao2026heracles,chen2026holomotion,qi2026humanoidgpt}, yet their best results depend on embodied motion corpora of hundreds of millions to billions of frames~\citep{luo2025sonic,qi2026humanoidgpt}.
Scaling the breadth and quality of tracker supervision has thus become a central practical bottleneck for zero-shot humanoid tracking.

Direct physical-robot demonstrations scale poorly: hardware rollouts are slow, safety-constrained, and expensive, whereas human motion is abundant.
Existing pipelines therefore acquire motion from mocap, video, or teleoperation, retarget it to the robot, and train tracking policies in simulation~\citep{he2024humanplus,he2024omnih2o,cheng2024asap,tessler2025protomotions,luo2025sonic,qi2026humanoidgpt}.
This enables large robot-space corpora, but the scalable substitute for robot demonstrations remains derived from human motion.
Hardware collection also undersamples the long-tail action compositions and
styles needed by a generalist, so its limitation is both cost and coverage.

\begin{figure*}[t]
  \centering
  \begingroup
  \definecolor{gapreference}{HTML}{D28E27}
  \definecolor{gapexecution}{HTML}{2F8E91}
  {\small
    \textcolor{gapreference}{\large$\bullet$}\,Reference
    \hspace{1.2em}
    \textcolor{gapexecution}{\large$\bullet$}\,Execution
  }\par
  \begin{tabular}{@{}c@{\hspace{0.02\textwidth}}c@{}}
    \textbf{(a) Shared motion feature space} &
    \textbf{(b) Physical descriptor distributions} \\
    \includegraphics[width=0.48\textwidth]{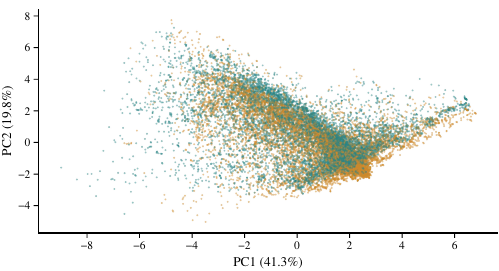} &
    \includegraphics[width=0.48\textwidth]{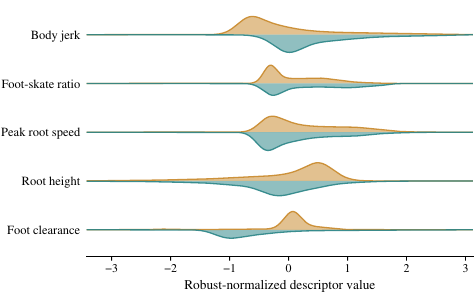}
  \end{tabular}
  \endgroup
  \caption{Reference--execution gap in diagnostic high-fidelity robot rollouts.
  \textbf{(a)} Shared PCA of 9,332 paired windows shows broad overlap with
  systematic local shifts. \textbf{(b)} Mirrored densities over all 10,369
  paired windows reveal changes in body jerk, foot skating, root height, and foot
  clearance, with a smaller change in peak root speed. Subset construction,
  normalization, and separability diagnostics are reported in the appendix.}
  \label{fig:distribution-gap}
\end{figure*}

The limitation is not purely a failure of retargeting.
Modern retargeters substantially improve kinematic correspondence and downstream tracking~\citep{araujo2025gmr}, yet expressing a motion in robot coordinates does not certify faithful closed-loop execution.
Residual mismatch in contacts, joint continuity, self-collision, or fast transitions persists, motivating kinodynamic and physics-aware refinement~\citep{chen2025ikretarget,zhao2026nmr}.
Hence, the scalable data route introduces two coupled bottlenecks: obtaining sufficient embodied coverage is costly, and the resulting human-retargeted corpus is only a proxy for the robot's executable motion distribution (Fig.~\ref{fig:distribution-gap}).
Crucially, validity is sequence- and controller-dependent: a clip can be well
formed frame by frame yet accumulate tracking error or fail at a contact
transition under closed-loop dynamics.

Pretrained motion generators offer a complementary coverage source.
Modern text-to-motion models already capture diverse actions, styles, and compositions~\citep{guo2022ml3d,tevet2023mdm,chen2023motiongpt,zhang2023t2mgpt,chen2023mld,pinyoanuntapong2024momask}, allowing diverse tracker references to be synthesized without additional robot demonstrations.
However, a generator reproduces its training distribution; when that distribution is human motion or its retargeted counterpart, sampling more motions merely expands the same proxy without improving execution compatibility.
It scales useful supervision only when samples are evaluated in robot space,
rather than assumed executable because their representation is robot-native.

\begin{figure*}[t]
  \centering
  \includegraphics[width=\textwidth]{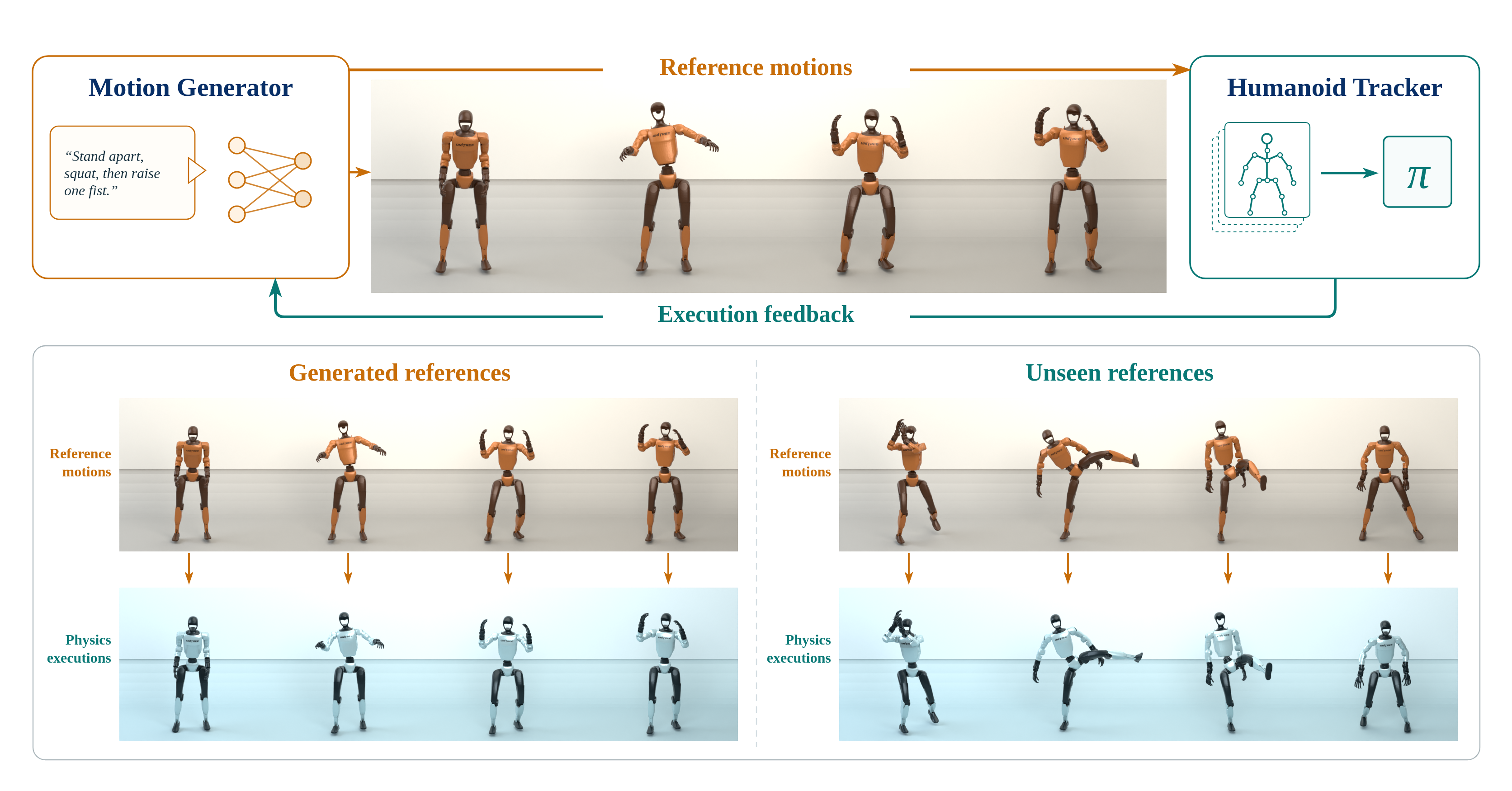}
  \caption{\textbf{Online generator--tracker co-training in robot space.}
  Generated robot-space references expand tracker supervision; their physics
  executions provide generator-alignment feedback.  At evaluation, the
  tracker executes held-out references.}
  \label{fig:teaser}
\end{figure*}

Existing work connects generation and control in one direction while fixing the other: generated motions can be tracked by downstream physics policies~\citep{rempe2026kimodo}, frozen generative priors can supervise policy training~\citep{zhang2025gmp}, and fixed trackers provide physical feedback for generator post-training~\citep{yue2025rlpf,zhang2026physmodpo}.
These formulations leave open a central question: can a pretrained generator reduce the embodied data needed to extend a zero-shot tracker's coverage while the tracker's feedback improves the generated distribution?
A fixed generated corpus cannot adapt as the tracker improves; one-shot tracker-filtered SFT aligns the generator to a frozen tracker, freezing the reward model and downstream tracker, thus failing to exploit later-executable motions or test whether generator supervision improves the tracker.
This motivates a joint online post-training loop where generated supervision and execution-grounded feedback co-evolve from available checkpoints.
The limitations are coupled: tracker improvements move the execution frontier
and make an offline-selected corpus stale, while generator updates expose
motions and failure modes absent from the fixed tracker-side pool.

We introduce \ours{}, an online generator--tracker framework that closes this loop (Fig.~\ref{fig:teaser}).
Starting from a pretrained robot-space motion generator and humanoid tracker, it alternates between synthesizing diverse references, evaluating them through closed-loop execution, aligning the generator with group-relative execution rewards, and updating the tracker with a mixture of generated and real retargeted data.
Generation expands tracker supervision beyond the available post-training corpus, while execution feedback guides the generator toward robot-compatible motions.
Both components evolve online, so the generated curriculum and its execution feedback co-adapt.
Generator collapse and tracker forgetting are constrained by regularizing toward the initial generator, supervised text–motion rehearsal, and real-reference replay.
These two branches address physical alignment and tracker coverage without using the current trainee as its own reward judge.

We validate \ours{} on two publicly released humanoid trackers with distinct pretraining paradigms, ProtoMotions and SONIC, to test generalization across backbone designs.
For each, we compare the pretrained checkpoint against equal-budget reference-only continuation, offline generator replay, and one-way component updates, isolating the effect of online mutual post-training.
Evaluation uses three frozen zero-shot tracking splits (LAFAN1-G1, AMASS-test-G1, Wild-G1-clean) and a private out-of-distribution generator test set; a common 30-FPS evaluator measures tracking success and trajectory errors, while official frozen-SONIC IsaacLab rollouts and TMR-G1 assess generator executability and semantic preservation.
The online co-training strategy consistently produces generators with higher physical fidelity and trackers with broader zero-shot coverage.
The SONIC-aligned branch yields broad tracking accuracy gains while preserving velocity fidelity; the ProtoMotions branch achieves improved out-of-distribution coverage and key-body accuracy, with split-dependent trade-offs.
These improvements are not reproduced by static replay or one-way filtering, indicating that closed-loop mutual adaptation drives the observed gains.

Our contributions are:
\begin{itemize}
  \item We identify that the two dominant bottlenecks in scaling zero-shot humanoid tracking---the cost of extending embodied supervision and the residual executability gap of retargeted references---are coupled and reinforce each other when addressed separately, motivating their joint treatment through closed-loop co-training.
  \item We propose \ours{}, an online framework that couples a text-to-motion generator and a humanoid tracker in a mutually improving loop: execution-grounded feedback aligns the generator toward robot-compatible motions, while newly generated references expand tracker coverage under controlled drift constraints.
  \item Across two distinct tracker backbones, controlled experiments show that online co-training improves both generator physical fidelity and tracker zero-shot coverage, with gains not reproduced by static replay or one-way filtering. The evaluation protocol isolates the effect of mutual adaptation from additional data, optimizer steps, and offline generation.
\end{itemize}

\section{Related Work}
\label{sec:related-work}

\paragraph{Physical feedback and closed-loop generation.}
Physics-aware generation reduces foot skating, penetration, unstable contacts,
and dynamically infeasible poses.  PhysDiff and CLoSD use simulation or
controller loops, while RobotMDM, RLPF, Humanoid-R0, and PhysMoDPO optimize
generators using a tracker-return surrogate, physical feedback,
deployment-oriented rewards, or execution preferences
\citep{yuan2024physdiff,tevet2025closd,serifi2024robotmdm,yue2025rlpf,
zhuang2025humanoidr0,zhang2026physmodpo}.
RoboForge and iterative closed-loop synthesis additionally feed optimized or
generated motions back into generator/control training
\citep{yuan2026roboforge,xu2026closedloop}.  Most closely, PARC alternates
character-motion generation with physics-based tracking correction, QuadFM
jointly trains a text-conditioned generator and controller for quadrupeds, and
Humanoid-DART iteratively updates a goal-conditioned humanoid trajectory
generator and tracker; a complementary humanoid locomotion system fine-tunes
the tracker against a frozen generator \citep{xu2025parc,gao2026quadfm,
debbad2026humanoiddart,zhang2026wholebodylocomotion}.  Thus, jointly updating
generation and control is not by itself our novelty.  \ours{} instead tests
whether tracker-derived feedback aligns a broad language-conditioned robot
motion generator and whether its online distribution improves unseen-motion
tracker generalization, against one-way, frozen-generator, and offline controls.

\paragraph{Baselines used in our study.}
For generalist tracking, we compare Any2Track's cross-embodiment policy,
Humanoid-GPT's language-conditioned controller, the AMP/PPO-based
ProtoMotions tracker, and the released SONIC whole-body tracker
\citep{zhang2025any2track,qi2026humanoidgpt,tessler2025protomotions,
luo2025sonic}.  BeyondMimic is a complementary per-reference specialist rather
than a zero-shot generalist \citep{liao2025beyondmimic}.  Matched
post-training controls comprise reference-only continuation, replay from the
frozen initial generator, and replay from the final online generator.  On the
generator side, we compare the initial robot-native model with one-way
tracker-filtered SFT, FlowGRPO against a frozen strong tracker, and the two
bidirectional variants.  Section~\ref{sec:experiments} specifies the matched
budgets and evaluation protocol for these baselines.

The supplementary material reviews text-to-motion generation,
language-conditioned humanoids, retargeting, and tracker-only systems in
greater detail.

\begin{table*}[t]
  \centering
  {\small
  \setlength{\tabcolsep}{2.2pt}
  \begin{tabular*}{\textwidth}{@{\extracolsep{\fill}}p{0.30\textwidth}ccccccc@{}}
    \toprule
    & \multicolumn{3}{c}{Fall-only SR (\%) $\uparrow$}
    & \multicolumn{4}{c}{All-trajectory error $\downarrow$} \\
    \cmidrule(lr){2-4}\cmidrule(lr){5-8}
    Method &
    LAFAN1 &
    AMASS-test &
    Wild-G1 &
    \metrichead{MPJPE\\(mm)} &
    \metrichead{$E_g$\\(mm)} &
    \metrichead{MPJVE\\(m/s)} &
    \metrichead{RootVelErr\\(m/s)} \\
    \midrule
    \multicolumn{8}{l}{\textit{Baselines}} \\
    Any2Track \citep{zhang2025any2track} & 100.0 & 5.1 & 10.4 & 320.9 & 1309.1 & 0.720 & 0.632 \\
    BeyondMimic \citep{liao2025beyondmimic} & 87.5 & 93.0 & 61.0 & 99.4 & 347.6 & 0.251 & 0.279 \\
    Humanoid-GPT \citep{qi2026humanoidgpt} & 85.0 & 83.3 & 71.4 & 128.1 & 1134.7 & 0.689 & 0.644 \\
    ProtoMotions ($T_0$) \citep{tessler2025protomotions} & 75.0 & 81.2 & 45.9 & 142.2 & 789.8 & 0.320 & 0.466 \\
    SONIC \citep{luo2025sonic} & 85.0 & 79.0 & 47.2 & 126.2 & 814.2 & 0.308 & 0.423 \\
    \midrule
    \multicolumn{8}{l}{\textit{ProtoMotions post-training}} \\
    Ref. only & 75.0 & 80.4 & 45.7 & 141.9 & 786.5 & 0.321 & 0.466 \\
    $G_0$ replay & \textbf{77.5} & 79.0 & 46.8 & 140.0 & \textbf{772.4} & \textbf{0.319} & \textbf{0.461} \\
    $G_{\rm final}$ replay & 77.5 & 78.3 & 46.4 & 140.7 & 756.2 & 0.318 & 0.456 \\
    \textbf{\ours{}} & 75.0 & \textbf{81.2} & \textbf{47.3} & \textbf{139.3} & 775.4 & 0.320 & 0.466 \\
    \midrule
    \multicolumn{8}{l}{\textit{SONIC post-training}} \\
    Ref. only & 82.5 & \textbf{79.7} & 45.9 & 131.8 & 867.6 & 0.314 & 0.436 \\
    $G_0$ replay & 85.0 & 78.3 & 45.5 & 133.3 & 847.6 & 0.321 & 0.433 \\
    $G_{\rm final}$ replay & 87.5 & 76.1 & 47.8 & 126.2 & 841.0 & 0.314 & 0.434 \\
    \textbf{\ours{}} & \textbf{90.0} & \textbf{79.7} & \textbf{48.0} & \textbf{124.1} & \textbf{807.2} & \textbf{0.308} & \textbf{0.423} \\
    \bottomrule
  \end{tabular*}}
  \caption{Zero-shot G1 tracking on three frozen splits under one fall-only
  protocol; errors pool all valid frames, including failures.  Within each
  backbone, the primary trainable rows share initialization and budget;
  Final-$G$ replay is an additional offline control.  Bold denotes the best
  primary trainable value per backbone.  Metric aggregation, provenance, and
  run contracts appear in Supplementary
  Section~\ref{app:evaluation-protocol}.}
  \label{tab:res-tracker-zeroshot}
\end{table*}

\begin{figure*}[t]
  \centering
  \includegraphics[width=\textwidth]{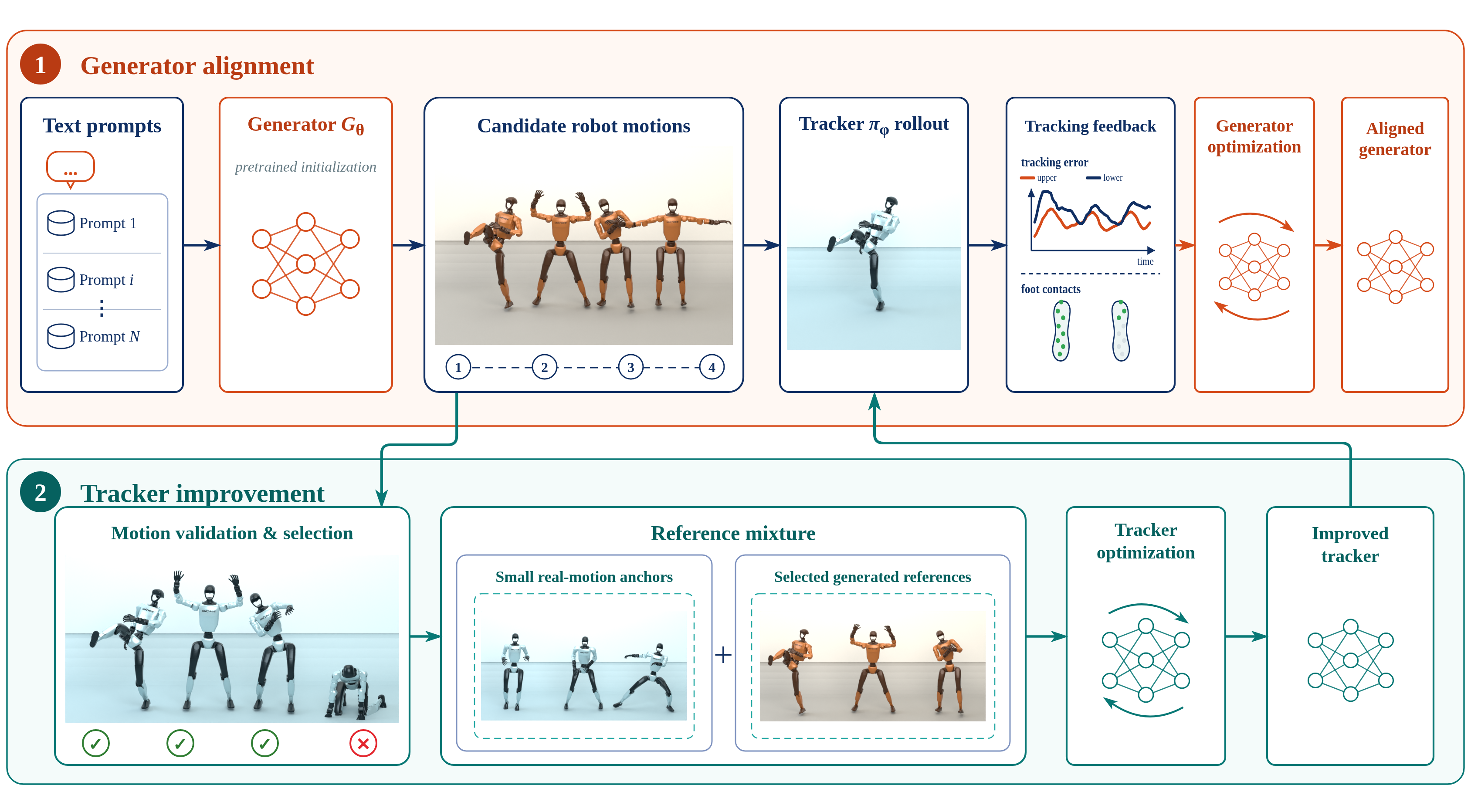}
  \caption{\textbf{\ours{} pipeline.}
  Starting from a pretrained generator and tracker, \ours{} alternates
  generator alignment via execution-grounded group-relative reward with tracker
  training on newly generated references.
  Both components evolve online; the tracker used for reward scoring is frozen
  within each generator phase and updated in the following round.}
  \label{fig:pipeline}
\end{figure*}
\section{Method}

\subsection{Problem formulation}

\paragraph{Motion generation.}
Let $\mathcal{M}_{R}$ denote the space of reference motions for a target robot.
A conditional motion generator models a distribution over motion sequences
$\robotmotion=(\robotmotion_1,\ldots,\robotmotion_T)\in\mathcal{M}_{R}$ given a
condition $\mathbf{y}$.  We write
$\robotmotion\sim p_{\theta}(\robotmotion\mid\mathbf{y})$, or equivalently
$\robotmotion=G_{\theta}(\mathbf{y},z)$ with stochastic input $z$.  The
condition may represent language or another motion specification; this work
studies text-conditioned generation, where $\mathbf{y}=\textprompt$.

\paragraph{Humanoid motion tracking.}
Given a reference $\robotmotion$, a tracker $\pi_{\phi}$ maps the current robot
state $s_t$ and a reference command $g_t(\robotmotion)$ to an action
$a_t\sim\pi_{\phi}(\cdot\mid s_t,g_t)$.  The robot dynamics produce a
closed-loop trajectory
$\boldsymbol{\tau}=\mathcal{E}(\pi_{\phi},\robotmotion)$.  Motion \mbox{tracking}
maximizes the expected discounted tracking return
\begin{equation}
  J_{\mathrm{track}}(\phi;\robotmotion)
  = \mathbb{E}\!\left[\sum_{t=1}^{T}\gamma^{t-1}
  r_{\mathrm{track}}(s_t,\robotmotion_t)\right].
  \label{eq:tracking-objective}
\end{equation}
Zero-shot tracking evaluates the same policy on reference motions not observed
during training, without motion-specific policy optimization.

\paragraph{Robot-native parameterization.}
Each frame uses a deterministic 38D G1 parameterization: a three-dimensional
root channel, continuous 6D pelvis rotation, and 29 actuated joints.  Per-clip
planar position and heading are canonicalized; planar motion is stored as
displacement while height remains absolute.  One-time offline GMD retargeting
supplies paired initialization data, but no retargeter runs inside the online
loop.  Supplementary Section~\ref{app:implementation} gives preprocessing
details.

\subsection{Framework overview}

\ours{} couples a pretrained motion generator and an existing humanoid tracker
in a bidirectional online training loop (Fig.~\ref{fig:pipeline}).  Starting
from a broad-coverage generator $G_{\theta_0}$ trained on large-scale
text--motion data and a pretrained humanoid tracker $\pi_{\phi_0}$, each round
samples diverse robot-space references, evaluates them through closed-loop
execution, and uses the resulting references and rollouts to update the two
models.  This sample--execute--update cycle provides the common backbone for
both directions of the framework.

The two models provide complementary supervision.  Structurally valid
on-policy generations accumulate as new tracker references, while tracker
updates draw equal numbers of public-reference and generated-reference
transitions.  Conversely, the tracker from the preceding round executes each
generated reference in closed loop; incomplete rollouts, joint and root
tracking errors, and unexpected falls form an execution score whose negative
provides the group-relative generator reward.  The scoring tracker remains
fixed within the generator phase and never gates tracker admission.  A
frozen-generator KL penalty and periodic supervised rehearsal on the original
text--motion data limit drift.  Alternating these updates lets the generated
pool and its execution reward evolve rather than fixing both offline.

\begin{table*}[t]
  \centering
  {\small
  \setlength{\tabcolsep}{2.0pt}
  \begin{tabular*}{\textwidth}{@{\extracolsep{\fill}}p{0.24\textwidth}ccccccc@{}}
    \toprule
    Generator &
    Succ. $\uparrow$ &
    $E_\mathrm{joint}\downarrow$ &
    $E_\mathrm{key}\downarrow$ &
    \metrichead{TMR\\R@1/2/3 $\uparrow$} &
    MM-Dist $\downarrow$ &
    FID $\downarrow$ &
    Diversity \\
    \midrule
    $G_0$ &
    92.58 & 0.159 & 0.410 &
    0.774/0.888/\textbf{0.931} & \textbf{18.293} & 0.023 & 36.375 \\
    Filtered SFT &
    \textbf{96.97} & \textbf{0.149} & 0.348 &
    0.771/0.886/0.929 & 18.450 & 0.028 & 36.264 \\
    Frozen tracker reward &
    90.92 & 0.158 & 0.363 &
    0.767/0.875/0.920 & 18.612 & 0.027 & 36.433 \\
    \textbf{\ours{} (ProtoMotions)} &
    93.55 & 0.160 & 0.399 &
    0.782/0.889/0.930 & 18.299 & \textbf{0.020} & 36.373 \\
    \textbf{\ours{} (SONIC)} &
    94.43 & 0.152 & \textbf{0.325} &
    \textbf{0.783}/\textbf{0.890}/0.930 & 18.302 & \textbf{0.020} & 36.371 \\
    \bottomrule
  \end{tabular*}}
  \caption{Generator post-training on a fixed 1,024-case suite under matched
  prompts, noise, decoding, and evaluation.  Frozen-SONIC rollouts measure
  physical quality; TMR-G1 measures semantic/distribution preservation.
  Bold marks the best value per metric (R-Precision componentwise, including
  ties); Diversity is descriptive.  Full definitions appear in Supplementary
  Section~\ref{app:metric-computation}.}
  \label{tab:res-generator-physics-style}
\end{table*}

\subsection{Generated-reference curriculum for tracker optimization}

Generated references extend an existing tracker without requiring a new large
tracker-specific reference corpus.  Let $\dataset_{\mathrm{base}}$
be a fixed seed reference pool, and let $\mathcal{P}_{\mathrm{train}}$ be a
fixed pool of training prompts.  The generator samples robot motions
$\dataset_{\mathrm{gen}}=\{\tgen(\textprompt_i,z_j)\mid \textprompt_i\in\mathcal{P}_{\mathrm{train}}\}$, and the tracker is trained on
\begin{equation}
  \dataset_{\mathrm{track}} =
  \dataset_{\mathrm{base}} \cup \dataset_{\mathrm{gen-valid}},
\end{equation}
where $\dataset_{\mathrm{gen-valid}}$ contains generated references that pass
reference-only structural checks: decoding must be finite and well formed, and
the motion must not require unavailable scene geometry.  We do not gate this
pool by rollout completion, velocity, amplitude, or current-trainee
difficulty.  Instead, every tracker update draws equal numbers of transitions
from $\dataset_{\mathrm{base}}$ and the accumulated
$\dataset_{\mathrm{gen-valid}}$, preventing pool growth from changing the
source mixture.

\subsection{Execution reward for generator alignment}

Tracker rollouts turn robot execution into a learning signal for text-to-motion
generation.  At round $r$, we freeze the tracker from the preceding round,
$\bar{\pi}^{(r)}$, throughout generator optimization and define
\begin{equation}
  R(\robotmotion;\bar{\pi}^{(r)}) =
  -S_{\mathrm{exec}}(\robotmotion;\bar{\pi}^{(r)}),
\end{equation}
where, for completion fraction $c$, maximum joint error $e_j$, mean
root-trajectory error $e_t$, and root-displacement error $e_d$,
\begin{equation}
  S_{\mathrm{exec}}=(1-c)+[e_j]_2+[e_t/0.5]_2
  +0.5[e_d/0.5]_2+2\mathbb{I}_{\mathrm{fall}},
\end{equation}
and $[x]_2=\min(x,2)$.  The current trainee has zero reward weight, and
velocity, amplitude, and binary-success gates are not used by the main
protocol.  Thus ``lagged'' denotes a stabilized in-branch reward model, not an
independent evaluator.

\subsection{Generator optimization}

Given $K$ samples from the same prompt, \ours{} updates the generator with a
group-relative FlowGRPO objective.  Rewards are normalized within each prompt
group, and sampled trajectories are replayed for multiple clipped policy-ratio
updates.  A frozen-generator KL anchor and periodic supervised flow-matching
updates on the original train-only text--motion pairs constrain drift from the
pretrained motion distribution
\citep{yue2025rlpf,liu2025flowgrpo}.  TMR-G1 retrieval and diversity are
held-out metrics and do not enter the reward.  The supplement specifies update
counts, anchor weights, replay versioning, and matched objective controls.

\subsection{Training schedule}

We realize \ours{} as an orchestrated alternation between the two branches.
Each round (1) samples $K$ same-prompt candidates and removes only invalid or
scene-dependent references; (2) scores the survivors with the frozen
preceding-round tracker and applies FlowGRPO with the two retention anchors;
(3) accumulates structurally valid on-policy generations and updates the
tracker with equal public/generated transition counts using its native
objective; and (4) exports that tracker for the next round.  We log reward
variance, policy-ratio movement, retention updates, exported references, and
frozen validation at every boundary.  The supplement gives initialization and
the complete schedule.

\begin{figure*}[!t]
  \centering
  \includegraphics[width=\textwidth]{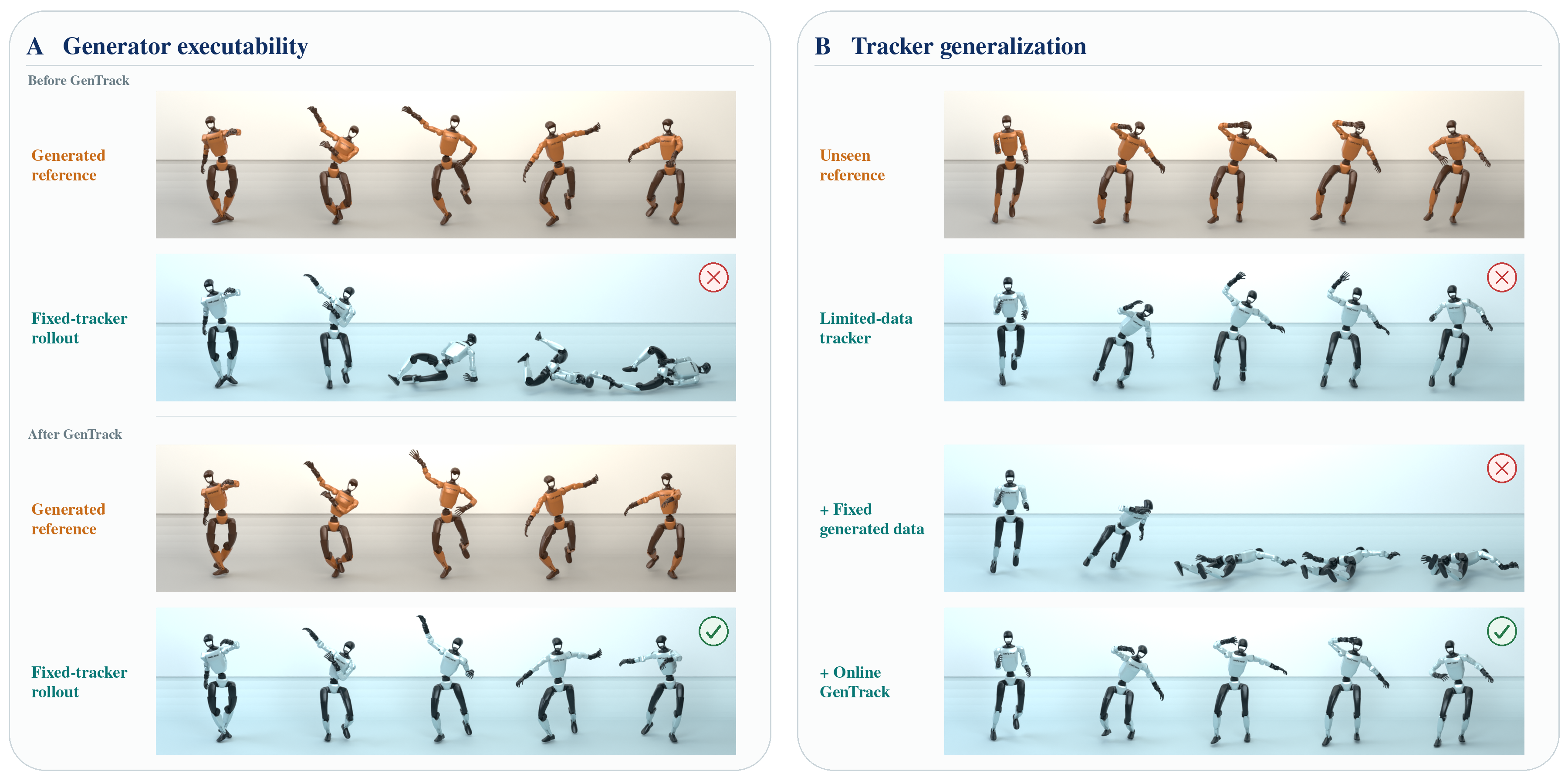}
  \caption{\textbf{Qualitative effects of bidirectional training.}
  \textbf{(a) Generator executability.} Given the same text prompt and sampling
  seed, a fixed SONIC tracker fails on the initial generated reference but
  successfully executes the reference produced by the GenTrack-aligned
  generator. \textbf{(b) Tracker generalization.} For the same held-out
  Wild-G1 reference, both the initial tracker and its frozen-$G_0$ replay
  counterpart fail, whereas the online-trained \ours{} tracker completes the
  motion. Each row shows the same five uniformly spaced phases from synchronized
  reference--execution trajectories.  For pose readability, every snapshot is
  independently pelvis-centered and yaw-normalized; this pose-normalized figure
  does not visualize the Table~\ref{tab:res-tracker-zeroshot} metric frame,
  global $E_g$, or the heading component of MPJPE.}
  \label{fig:qualitative-bidirectional}
\end{figure*}

\section{Experiments}
\label{sec:experiments}

We evaluate whether online generated references extend zero-shot tracker coverage beyond matched continuation and static replay, and whether closed-loop tracker feedback improves generator physical fidelity while preserving semantic quality. Controlled one-way baselines isolate each direction.

\subsection{Experimental Setup}

\paragraph{Protocol and data.}
All experiments use Unitree G1 and 30-FPS exports.  We instantiate \ours{}
from fixed ProtoMotions \citep{tessler2025protomotions} and SONIC
\citep{luo2025sonic} checkpoints; within each backbone, trainable rows share
initialization, simulator, optimizer, transition budget, validation, and test
manifests.  The tracker anchor pool contains 13,337 AMASS/LAFAN references.
Generator initialization instead uses 357,472 GMD-retargeted G1 pairs from an
internal human-motion corpus.  These pairs initialize and rehearse only the
generator; tracker post-training uses the AMASS/LAFAN pool and
generator-sampled references.
Evaluation uses frozen LAFAN1-G1, AMASS-test-G1, Wild-G1-clean, and a separate
1,024-prompt generator suite; no test item or metric enters training.
External rollouts receive the same post-hoc processing, and all generator rows
share one frozen SONIC executor.  Supplementary
Sections~\ref{app:implementation} and~\ref{app:evaluation-protocol} give
filtering, checkpoint provenance, split exclusions, and run contracts.
Offline controls also match accepted-motion counts or tracker transitions as
applicable; released native success flags are excluded from the common
comparison.

\paragraph{Metrics.}
We evaluate the tracker baselines introduced in
Section~\ref{sec:related-work} by common physical-coverage and motion-fidelity
measures, and evaluate generators by frozen-SONIC executability plus TMR-G1
semantic and distribution measures.  Full definitions, body sets, alignment,
thresholds, aggregation, and diagnostic-only rules are given in supplementary
Section~\ref{app:metric-computation} and
Table~\ref{tab:metric-definitions}.

\subsection{Main Tracker Post-Training Results}

Table~\ref{tab:res-tracker-zeroshot} compares external trackers with matched
post-training variants.  With ProtoMotions, static replay trades AMASS-test SR
for LAFAN1/Wild-G1 gains; \ours{} recovers AMASS-test, attains the best Wild-G1
SR, and reduces MPJPE from 142.2 to 139.3 mm, although its $E_g$ remains above
frozen-$G_0$ replay.  With SONIC, \ours{} raises SR from 85.0/79.0/47.2 to
90.0/79.7/48.0, reduces MPJPE from 126.2 to 124.1 mm and $E_g$ from 814.2 to
807.2 mm, and matches both velocity errors without replaying Wild-G1-clean.
Relative to frozen-$G_0$ replay, these changes improve success on all three
splits while lowering MPJPE by 9.2~mm and $E_g$ by 40.4~mm.  Final-$G$ replay
remains below the online SONIC row on every reported measure, showing that the
final generated pool alone does not reproduce the online training trajectory.
Final-$G$ replay provides an additional offline control; Supplementary
Section~\ref{app:evaluation-protocol} gives its run contract and the
external-baseline construction.

\subsection{Main Generator Results}

Table~\ref{tab:res-generator-physics-style} evaluates all generators under identical prompts, noise, lengths, and a frozen SONIC evaluator.
Tracker-filtered SFT uses a frozen ProtoMotions \( T_0 \) to filter candidates for supervised fine-tuning; SONIC is only the common post-hoc evaluator.
While SFT achieves high nominal success (96.97), this metric is misleading: it optimizes for conformity to a single, frozen tracker's biases rather than for genuine physical plausibility.
Consequently, SFT's generations collapse toward low-difficulty motions that \( T_0 \) already executes well, as evidenced by its degraded semantic metrics, distributional quality, and higher FID.
This highlights a fundamental limitation of one-way filtering: it creates a self-reinforcing loop that sacrifices motion diversity and difficulty for inflated success on a static judge.

In contrast, \ours{} (SONIC) demonstrates substantial physical improvement without this trade-off.
It reduces \( E_{\text{key}} \) from 0.410 to 0.325 m—surpassing SFT in key-body fidelity—while improving R@1/2 to 0.783/0.890, preserving R@3 at 0.930, and lowering FID to 0.020.
The ProtoMotions variant yields similar semantic gains.
Together, the physical and semantic metrics show that \ours{} improves robot
compatibility while preserving generation difficulty and distributional
quality relative to tracker-filtered SFT.

\subsection{Ablation Study}

The main tables already separate one-way controls: frozen-$G_0$ replay helps some splits but regresses AMASS-test, while tracker-filtered SFT improves executability at the cost of semantic preservation, confirming that neither static replay nor one-way alignment alone reproduces the full effect of mutual adaptation.
Generator-only training with the same execution reward and retention anchors
raises frozen-SONIC success from 92.58 to 93.95, reduces
$E_\mathrm{key}$ from 0.410 to 0.381, and changes R@1/FID from
0.774/0.023 to 0.783/0.020.  The
online SONIC branch further improves success and both physical errors,
supporting a contribution from bidirectional adaptation beyond generator-only
alignment (Supplementary Section~D).
Among matched objective controls, success-only feedback reaches 93.12
execution success with 0.780 R@1 and 0.021 FID, whereas using the current
trainee as sole judge lowers cross-split SR to 74.8/79.6/46.4.
Reward-weighted SFT attains 94.10 success and 0.158/0.389 joint/key-body error
but degrades R@1/FID to 0.769/0.029.  Diffusion-DPO is more balanced at 93.36
success, 0.776 R@1, and 0.023 FID, but remains below the full loop on
AMASS-test/Wild-G1 coverage and generator success, key-body error, R@1, and
FID.  Matched anchor-weight controls are reported in the supplement.
Together, these controls indicate that the online loop yields benefits beyond
its individual components.

\subsection{Qualitative Analysis}

Figure~\ref{fig:qualitative-bidirectional} visualizes representative cases under fixed conditions: same prompt, noise, and length for the generator; a held-out Wild-G1 reference for the tracker.
Generator comparisons show \ours{} removing failure-inducing artifacts (e.g., abrupt joint reversals) that cause $G_0$ outputs to fail during execution.
Tracker comparisons illustrate recovery of difficult phases---deep crouches, rapid turns, contact transitions---where the pretrained tracker fails but the jointly updated tracker succeeds.
These examples illustrate the mechanisms behind the quantitative results in Tables~\ref{tab:res-tracker-zeroshot} and~\ref{tab:res-generator-physics-style}; selection criteria, counterexamples, and descriptor-space analysis are detailed in Supplementary Section~D.2.

\subsection{Diagnostic Scope and Limitations}

Aggregate tracking metrics do not reveal \emph{how} retargeted references and executed motions differ.
Figure~\ref{fig:distribution-gap} characterizes this residual gap (which motivates \ours{}) through shared PCA (first two components explain 61.1\% of pooled variance) and a motion-grouped classifier (AUC $0.996{\pm}0.004$), confirming the gap is systematic yet the domains share substantial structure.
Per-descriptor densities localize the mismatch to jerk, skating, root height, and foot clearance.
These results show that binary success under-characterizes the gap, while the overlap motivates online adaptation rather than discarding the retargeted corpus.
Supplementary Section~D.2 provides subset construction and normalization details.

\section{Conclusion}

We presented \ours{}, a bidirectional generator--tracker loop for simulated
Unitree G1.  Joint post-training improves five SONIC tracker metrics and
matches two velocity errors, while ProtoMotions yields trade-offs.  Against
$G_0$, the SONIC-aligned generator improves all three physical metrics;
against tracker-filtered SFT, it trades success and joint error for key-body
fidelity and text/distribution preservation.  Together, these results support
online co-training as an effective way to improve robot-compatible motion
generation and extend zero-shot tracking coverage across two tracker
backbones.

\bibliography{references}

\clearpage
\setcounter{secnumdepth}{2}
\setcounter{dbltopnumber}{1}
\renewcommand{\dbltopfraction}{0.90}
\renewcommand{\dblfloatpagefraction}{0.75}
\setlength{\dbltextfloatsep}{8pt plus 2pt minus 2pt}
\appendix
\section{Extended Related Work}
\label{app:related-work}

Because of the main-paper page limit, its Related Work section gives a concise
overview; this appendix provides a broader discussion of closely related work.

\paragraph{Text-to-motion generation.}
Text-to-motion generation maps natural language to human motion sequences and is
commonly evaluated on HumanML3D and KIT-ML with retrieval, FID, diversity, and
multimodality metrics \citep{plappert2016kit,guo2022ml3d}.  Diffusion
models such as MDM, MotionDiffuse, MLD, ReMoDiffuse, FlowMDM, and MotionLCM
improve quality, controllability, retrieval augmentation, long-horizon
composition, or sampling efficiency \citep{tevet2023mdm,zhang2022motiondiffuse,
chen2023mld,zhang2023remodiffuse,barquero2024flowmdm,dai2024motionlcm}.
Discrete-token and language-model variants such as T2M-GPT, MotionGPT, MoMask,
MMM, SATO, and HumanTOMATO improve sequence modeling, text alignment, and
whole-body expressiveness \citep{zhang2023t2mgpt,chen2023motiongpt,
pinyoanuntapong2024momask,pinyoanuntapong2024mmm,chen2024sato,
lu2024humantomato}.  Their standard benchmarks measure semantic fidelity and
motion diversity, but do not determine whether a generated reference is
executable by a physical humanoid.

\paragraph{Language-conditioned humanoid motion.}
Recent work has begun to connect motion generation with humanoid execution.
Kimodo scales controllable kinematic motion generation and releases
robot-compatible assets \citep{rempe2026kimodo}.  TextOp, FRoM-W1,
Humanoid-LLA, SafeFlow, MIND, and SCRIPT drive humanoids from language by
combining high-level motion or intent generation with physics-based execution
or action policies \citep{xie2026textop,li2026fromw1,liu2025humanoidlla,
cho2026safeflow,li2026mind,zhang2026script}.  DreamControl-v2, PhyGile, CLAW,
and RoboGhost further study robot-space priors, physics-prefix generation,
language-annotated G1 data construction, or retargeting-free language-to-control
interfaces \citep{harithas2026dreamcontrolv2,bao2026phygile,cao2026claw,
li2026roboghost}.  \ours{} is closest to this emerging family, but targets a
different experimental claim: a robot-native generator should improve under
physical feedback, and the tracker should improve zero-shot coverage from the
generated distribution rather than merely execute the generator's current
outputs.

\paragraph{Joint generator--controller training.}
RobotMDM learns a motion-only critic that predicts the expected return of a
frozen tracker, uses it as a differentiable loss to fine-tune a
text-conditioned robot-motion diffusion model, and separately retrains trackers
on generated motions \citep{serifi2024robotmdm}.
PARC iteratively retrains a character motion generator on trajectories corrected
by a physics-based tracker, while QuadFM jointly trains text-conditioned motion
generation and control for quadrupeds \citep{xu2025parc,gao2026quadfm}.
Humanoid-DART applies an alternating generator--tracker curriculum to sparse,
goal-conditioned humanoid loco-manipulation demonstrations
\citep{debbad2026humanoiddart}.  Other generator--tracker hierarchies improve a
humanoid tracker online while freezing the generator
\citep{zhang2026wholebodylocomotion}.  These are direct precedents for coupling
the two components.  Our scope differs in testing broad language-conditioned
robot motion, bidirectional post-training, and unseen-motion tracker coverage as
separate claims under matched one-way and offline controls.

\paragraph{Retargeting and morphology.}
Human-to-humanoid retargeting maps SMPL, marker, video, or BVH motions to robot
joint trajectories under morphology, contact, and joint-limit constraints.  H2O
and OmniH2O use retargeted human motion to train real-time humanoid
teleoperation policies \citep{he2024h2o,he2024omnih2o}.  Recent retargeting
studies such as General Motion Retargeting, implicit kinodynamic retargeting,
kinodynamic trajectory optimization, physics-aware cross-morphology transfer,
and neural motion retargeting show that retargeting quality strongly affects
downstream tracking and controllability \citep{araujo2025gmr,
chen2025ikretarget,zhang2026kdmr,huang2026human2humanoid,zhao2026nmr}.
In our experiments, retargeting is therefore
not treated as an invisible preprocessing step: native G1 generation,
held-out GMD-retargeted G1 references, and tracker execution failures are measured
separately.

\section{Implementation Details}
\label{app:implementation}

\subsection{Robot Choice and Motion Representation}

We use Unitree G1 as the primary robot morphology because it is an open humanoid robot model with ProtoMotions support, MJCF/URDF assets, pretrained tracker checkpoints, and a real-robot deployment path.  This choice makes the robot-control claim concrete: success is measured on an actual humanoid morphology rather than on an SMPL-like physics proxy.

\paragraph{Robot-native motion representation.}
The generator emits a compact per-frame representation $\robotmotion_t \in \mathbb{R}^{38}$ that decodes exactly to a MuJoCo G1 configuration: a $3$-d root channel, a $6$-d continuous rotation \citep{zhou2019rot6d} of the pelvis orientation, and the $29$ actuated joint angles.  Two design choices make this target a deterministic function of language and easy for a flow-matching model to fit.  \emph{(i) Per-clip canonicalization.}  The raw retargeted clips inherit an arbitrary world placement: the ground-plane $(x,y)$ start position and the global heading (yaw about the up axis) are not determined by the caption, so a text-conditioned model regresses them toward the dataset mean and incurs metre-scale translation error.  We therefore express every clip in a canonical frame whose first frame sits at the ground origin facing the canonical $+x$ direction, while preserving the physically meaningful height channel.  \emph{(ii) Velocity-based root.}  The ground-plane translation is stored as a per-frame displacement rather than an absolute position; absolute position spans a large, sparse range (a run reaches several metres) that a flow-matching model systematically under-shoots, whereas the per-frame velocity is small and bounded, and the absolute trajectory is recovered by cumulative summation at decode time.  The height channel stays absolute because it is bounded and semantically informative (e.g.\ a crawl versus a stand).

\paragraph{Offline retargeting as data construction.}
To obtain robot-native supervision, we retarget our internal human-motion dataset to the Unitree G1 skeleton \emph{once, offline}, using the GMD retargeting pipeline.  After internal quality control, this yields a paired training set $\dataset_{\mathrm{pair}}=\{(\textprompt_i, \robotmotion_i)\}$ with 357,472 clips.  We use these internal pairs only to initialize and rehearse $\tgen$; their GMD-retargeted trajectories are never supplied to tracker training.  We warm-start $\tgen$ from a pre-trained human text-to-motion backbone, replace its output head with the $38$-d robot head, and supervise it on $\dataset_{\mathrm{pair}}$.  Crucially, retargeting is confined to this offline generator-training stage: at loop time the generator already speaks the robot's language, which removes the retargeting confound that otherwise contaminates any claim about generator physical realism.

\subsection{Generator Retention and Initialization}

Physical reward alone can favor slow, low-amplitude motions that are easy to
execute.  The full configuration limits drift with two retention
mechanisms: a KL penalty to the frozen initial generator during FlowGRPO and
periodic supervised flow-matching updates on the original train-only
text--motion pairs.  Generator prompts are sampled from the fixed curated
training corpus.  No TMR, style, diversity, or prompt-coverage reward enters the
training objective.  Instead, TMR-G1 retrieval, MM-Dist, FID, and diversity are
measured on a frozen held-out suite for model assessment.

\paragraph{Offline initialization.}
We build $\dataset_{\mathrm{pair}}$ by retargeting our internal human-motion
dataset to the G1 skeleton with GMD and fine-tune the robot-native generator
$\tgen$ on it.  These pairs are used only for generator initialization and
rehearsal and never enter tracker training.
The tracker $\tracker^{(0)}$ is initialized from an existing pretrained policy.
We instantiate the framework separately with pretrained ProtoMotions and
released SONIC trackers; their histories are disclosed and never counted as
new GenTrack supervision.  The initial tracker supplies the first quality
model, after which each round uses the tracker from the preceding round.

\subsection{Tracker Curriculum and Generator Update Details}

Generator prompts undergo an internal validity check before generation.  The
main protocol rejects only malformed or non-finite decoded
references; it does not require the lagged quality tracker or current trainee
to complete a motion before admitting that reference to tracker training.
Tracker updates draw equal numbers of public-reference and generated-reference
transitions.  Generated-motion and environment-transition budgets are matched
to the offline controls.

Let $c$ denote rollout completion, $e_j$ maximum joint error in radians,
$e_t$ mean root-trajectory error in metres, and $e_d$ root-displacement error
in metres.  The implemented execution score is
\begin{equation}
  \begin{aligned}
    S_{\mathrm{exec}}={}&(1-c)+[e_j]_{2}+[e_t/0.5]_{2}\\
    &+0.5[e_d/0.5]_{2}+2\mathbb{I}_{\mathrm{fall}},
  \end{aligned}
\end{equation}
where $[x]_2=\min(x,2)$.  The quality tracker from the preceding round remains
fixed throughout each generator phase, while the current trainee has zero
generator-reward weight.  No additional velocity or amplitude gate is applied
to this score.

For FlowGRPO, rewards are normalized within each same-prompt sample group and
each freshly sampled trajectory group is replayed for four clipped policy-ratio
updates.  The full configuration uses a $0.02$ KL weight to the frozen generator
and a supervised flow-matching anchor update of weight $1.0$ every two GRPO
iterations.
The reward-weighted SFT control uses the same positive-advantage samples, while
the Diffusion-DPO control forms preferences from the highest- and lowest-reward
samples in each prompt group.  Sampling and update budgets are matched across
all three objectives.

\paragraph{Online co-training schedule.}
Each round first samples $K$ candidates per training prompt.  Reference-only
checks remove malformed or scene-dependent motions, and the lagged quality
tracker scores the survivors.  Same-prompt rewards then create FlowGRPO
advantages, and multi-epoch replay updates the generator under KL and GT
anchors.  Structurally valid on-policy generations accumulate in a merged pool
with the fixed public references, and the tracker is updated by its
method-native objective with an equal public/generated transition budget.  The updated tracker
becomes the next trainee, while the previous trainee is delayed into the quality
tracker role.  Reward variance, policy-ratio movement, KL and GT-anchor updates,
exported references, and frozen validation performance are checked at
every round boundary.

\section{Detailed Experimental Protocol}
\label{app:evaluation-protocol}

\begin{table*}[!t]
  \centering
  {\small
  \setlength{\tabcolsep}{3.5pt}
  \begin{tabular*}{\textwidth}{@{\extracolsep{\fill}}p{0.22\textwidth}ccccccc@{}}
    \toprule
    \textbf{Source} &
    \metrichead{\textbf{SONIC}\\\textbf{Succ. $\uparrow$}} &
    \metrichead{\textbf{Root}\\\textbf{err. $\downarrow$}} &
    \metrichead{\textbf{Ref.}\\\textbf{skate $\downarrow$}} &
    \metrichead{\textbf{TMR}\\\textbf{R@1 $\uparrow$}} &
    \textbf{MM-Dist $\downarrow$} &
    \textbf{FID $\downarrow$} &
    \textbf{Diversity} \\
    \midrule
    Retargeted GT &
    91.41 & 0.402 & 0.126 &
    \textbf{0.849} & \textbf{16.345} & \textbf{0.000} & 36.713 \\
    KIMODO-G1 \citep{rempe2026kimodo} &
    \textbf{98.63} & \textbf{0.101} & \textbf{0.087} &
    0.500 & 25.215 & 0.122 & 35.689 \\
    HYMotion ($G_0$) &
    91.89 & 0.381 & 0.119 &
    0.774 & 18.293 & 0.023 & 36.375 \\
    \bottomrule
  \end{tabular*}}
  \caption{Robot-native source validation on the fixed held-out test suite.
  SONIC Succ. and Root err. are measured by executing each reference with the
  same frozen IsaacLab SONIC policy.  Succ. is recomputed from complete 30-FPS
  exports with the shared fall-only protocol; released callback flags are
  diagnostic only. Ref. skate is computed on each reference with the same
  MuJoCo-FK contact protocol. Semantic metrics use TMR-G1, with MM-Dist and
  Diversity in raw latent space and FID in unit-normalized latent space. Succ.
  is a percentage, Root err. is in metres, and Ref. skate is in metres per
  second. Bold marks the best value per metric; Diversity is descriptive.}
  \label{tab:generator-backbone-selection}
\end{table*}

\begin{table*}[!t]
  \centering
  {\footnotesize
  \setlength{\tabcolsep}{3pt}
  \renewcommand{\arraystretch}{0.95}
  \begin{tabular*}{\textwidth}{@{\extracolsep{\fill}}p{0.17\textwidth}p{0.49\textwidth}p{0.20\textwidth}c@{}}
    \toprule
    \textbf{Metric} & \textbf{Definition} & \textbf{Used for} &
    \textbf{Direction} \\
    \midrule
    Generator Succ. &
    fraction of generated references whose complete 30-FPS official SONIC execution maintains reference-relative pelvis-height deviation at most $0.25$ m \citep{luo2025sonic,yue2025rlpf} &
    generator physical quality & $\uparrow$ \\
    $E_\mathrm{joint}$ &
    mean wrapped rotation error over the 29 actuated G1 joints between reference and execution, in radians &
    generator physical quality & $\downarrow$ \\
    $E_\mathrm{key}$ &
    mean start-XY-aligned position error over SONIC's fixed official 14 tracked bodies, in metres &
    generator physical quality & $\downarrow$ \\
    Fall-only SR &
    fraction of unseen references whose complete 30-FPS rollout maintains maximum reference-relative pelvis-height deviation at most $0.25$ m; one missing resampled endpoint is tolerated \citep{luo2025sonic} &
    tracker zero-shot coverage & $\uparrow$ \\
    Completion / Unexpected Fall &
    executed-frame ratio and persistent reference-relative collapse when reference pelvis height and up-axis cosine are at least $0.50$; low-floor targets are excluded and absolute execution height is never used &
    auxiliary diagnostics & mixed \\
    MPJPE &
    mean Cartesian position error over SONIC's official 14-body set after subtracting each frame's pelvis translation, in millimetres; all cases and frames pooled across all splits in one frame-micro average \citep{luo2025sonic} &
    tracker pose fidelity & $\downarrow$ \\
    $E_g$ &
    start-XY-aligned global Cartesian position error over the same 14-body set, in millimetres; the same all-trajectory aggregation as MPJPE &
    tracker global fidelity & $\downarrow$ \\
    MPJVE &
    mean finite-difference Cartesian velocity error over the same root-relative 14-body trajectories, in metres per second; the same all-trajectory aggregation as MPJPE &
    tracker motion fidelity & $\downarrow$ \\
    RootVelErr &
    mean Cartesian linear-velocity error of the robot root, in metres per second; the same all-trajectory aggregation as MPJPE &
    tracker trajectory fidelity & $\downarrow$ \\
    Root err. &
    mean start-aligned root-trajectory position error between reference and execution &
    physical feasibility & $\downarrow$ \\
    Foot skate / penetration &
    contact-frame foot speed and below-ground penetration depth computed on generated references or physics rollouts &
    physical artifacts & $\downarrow$ \\
    Robot kinematic diagnostics &
    joint-limit saturation, contact-rhythm mismatch, root/yaw statistics, and other robot-centric pose or velocity checks; these are auxiliary diagnostics rather than main distribution metrics &
    morphology compatibility audit & mixed \\
    Prompt cov. / action entropy &
    prompt/action coverage and action-category entropy over the fixed generator prompt suite &
    anti-degeneration & $\uparrow$ \\
    Judge calibration &
    margin by which lagged checkpoints from the adapting tracker branch rank held-out GT references above corrupted or invalid references before their reward is used for generator optimization &
    reward reliability & $\uparrow$ \\
    Text guardrails &
    TMR-G1 R-Precision and raw-latent MM-Dist/diversity, together with unit-normalized-latent FID, on the fixed robot-motion suite &
    semantic preservation & mixed \\
    \bottomrule
  \end{tabular*}}
  \caption{Metric definitions and evaluation roles.  Generator metrics evaluate
  text-conditioned references with frozen-SONIC execution and TMR-G1, while
  tracker metrics evaluate policies executing fixed references.  Low-level
  generator metrics follow the RLPF structure, and tracker success follows
  SONIC's relaxed criterion recomputed from shared 30-FPS G1 trajectories.}
  \label{tab:metric-definitions}
\end{table*}

\paragraph{Robot and motion representation.}
All experiments use the Unitree G1 morphology and the canonical $38$-d motion
representation described above.  References, generated motions,
executed trajectories, and visualizations are exported at 30 FPS.  Matched
training comparisons use the same tracker architecture, simulator, optimizer,
initialization, and environment-transition budget.  External trackers are run
with their released implementations, but their outputs are converted to the
same start-aligned G1 reference/execution format before root-aware tracker
evaluation.  Their native success flags are not used for tracker comparison.
Generator executability is evaluated separately with a frozen SONIC policy and
the low-level metric family defined below.  This evaluator is shared by every
generator row and provides a consistent compatibility measure under frozen
SONIC execution on simulated G1.

\paragraph{Training data.}
The generator-only internal corpus contains 357,472 GMD-retargeted G1 training
pairs and a fixed held-out test set.  We use disjoint training and
test partitions and apply an internal curation and quality-control protocol.
Across tracker post-training arms, the only admissible sources are the public
pool of 12,733 AMASS-G1 and 604 LAFAN1-G1 motions and G1 references sampled from
the relevant generator; reference-only continuation uses the public pool alone.
Captions from the internal training split may provide generator prompts, but
their paired GMD-retargeted trajectories are never provided to the tracker.

\paragraph{Tracker protocols and baselines.}
We instantiate \ours{} from two existing tracker checkpoints: a
root-aware ProtoMotions/AMP-PPO checkpoint
\citep{tessler2025protomotions} and the official released SONIC checkpoint
\citep{luo2025sonic}.  For each backbone, the primary post-training arms start
from the same checkpoint hash and match simulator, optimizer, environment
transitions, validation, and checkpoint selection.  The primary comparison is
the frozen checkpoint versus equal-budget public-reference continuation,
frozen-$G_0$ offline replay, and \ours{}.  We additionally report offline replay
from the final online generator.  This control restarts from the same tracker
checkpoint and separates temporal co-adaptation from final-generator quality.
Checkpoint pretraining data are disclosed rather
than counted as evidence for training a tracker from scratch.
Released Any2Track \citep{zhang2025any2track}, Humanoid-GPT
\citep{qi2026humanoidgpt}, SONIC, and BeyondMimic
\citep{liao2025beyondmimic} are external references; their weights are not used
to initialize either adapting branch, except for the explicitly labeled
SONIC-Released branch itself.

\paragraph{Evaluation splits.}
We evaluate zero-shot motion generalization on three frozen, mutually
disjoint G1 splits.  LAFAN1-G1 follows the common locomotion and transition
benchmark \citep{harvey2020robust}; AMASS-test-G1 follows the protocol used by
prior tracking work \citep{mahmood2019amass,tessler2025protomotions}; and
Wild-G1-clean is held out from the robot-native corpus and curated for
compatibility with the flat-ground evaluation setup.  No test split is used for
training, replay construction, or checkpoint selection.

\subsection{Metric Definitions and Computation}
\label{app:metric-computation}

\paragraph{Tracker metrics.}
Every tracker is first rolled out until the reference motion timeout with its
method-native failure termination disabled.  We export reference and execution
states, resample each trajectory once to 30 FPS, and apply the same post-hoc
termination test to every method.  Following SONIC's cross-method MuJoCo
comparison \citep{luo2025sonic}, a trajectory fails only if its maximum
reference-relative pelvis-height error exceeds $0.25$ m.  A single missing
endpoint sample caused by resampling is tolerated.  Because the height test is
reference-conditioned, a commanded fall or get-up is not rejected merely for
approaching the ground.  The released-code evaluator's additional ankle/wrist
height and pelvis-orientation terms, method-native success flags, and
Any2Track's mean-error criterion are retained only as diagnostics and never
enter the main table.

All simulator evaluations start from reference frame zero.  We disable
observation corruption, reset pose/velocity perturbations, adaptive motion
sampling, and startup randomization of friction, restitution, default joint
positions, center of mass, and body mass.  These choices make the nominal
initial state deterministic across checkpoints.

Using SONIC's root-relative position definition and the position--velocity
metric family used by Humanoid-GPT \citep{luo2025sonic,qi2026humanoidgpt}, we
report mean per-joint position error (MPJPE), start-XY-aligned global joint
position error ($E_g$), mean per-joint velocity error (MPJVE), and root
linear-velocity error (RootVelErr).  Because the source papers use different
representations and units, all four are recomputed in Cartesian robot space
from the same synchronized trajectories.  MPJPE subtracts the pelvis
translation from each pose, evaluates SONIC's fixed 14-body set, and is
reported in millimetres.  $E_g$ retains global joint positions after start-XY
alignment and is reported separately in millimetres.  MPJVE applies the same
root-relative alignment and body set to finite-difference velocities (m/s),
while RootVelErr measures global pelvis linear-velocity error (m/s).

All four errors pool every valid frame from every evaluated case over the three
splits into one frame-micro average.  Split-macro and success-only values remain
diagnostics and never populate the main table.  Fall-only SR is shown alongside
the errors as the separate coverage statistic.  Completion, reference-conditioned
Unexpected Fall, and acceleration error remain auxiliary
diagnostics and do not enter the success decision or the main table.

\paragraph{Generator metrics.}
Following the high-level/low-level evaluation decomposition of RLPF
\citep{yue2025rlpf}, generator quality is measured by semantic generation
metrics and by official frozen-SONIC execution in IsaacLab.  We report success
rate $\mathrm{Succ}$, mean wrapped actuated-joint
rotation error $E_\mathrm{joint}$ (rad), and mean start-XY-aligned position
error $E_\mathrm{key}$ (m) over SONIC's fixed official 14 tracked bodies.  We
use clearer notation because RLPF's ``MPJPE (rad)'' is a joint-rotation rather
than a Cartesian position error.  $\mathrm{Succ}$ is
recomputed from complete 30-FPS exports using the same fall-only,
reference-conditioned termination term as the tracker table.  The frozen
policy, motion-timeout horizon, reference, and start alignment are held fixed.
Reported durations and continuous errors are computed from complete 30-FPS
exports rather than simulator-specific early-reset flags.  Unexpected Fall, root
error, foot skate, and penetration remain auxiliary diagnostics rather than
extra success gates.

TMR-G1 supplies R-Precision, MM-Dist, FID, and diversity.  MM-Dist and
diversity use raw TMR-G1 latents,
whereas FID uses unit-normalized latents to avoid scale-dominated Fr\'echet
values.  SMPL- or HumanML3D-space evaluators are not used for the main
generator result.

\subsection{Robot-Native Source Validation}

Table~\ref{tab:generator-backbone-selection} evaluates the held-out G1
references and KIMODO-G1 \citep{rempe2026kimodo} under the same prompts and
evaluators.  The frozen initial generator $G_0$ is reported in
main-paper Table 2 and is not repeated here.  The GT
physical columns are obtained by executing the retargeted references with the
frozen tracker, rather than by a trivial reference-to-reference comparison.
This experiment validates the motion source and execution judge used by the
subsequent closed-loop study.

\section{Additional Experiments and Ablations}
\label{app:additional-experiments}

\begin{table*}[!t]
  \centering
  {\small
  \setlength{\tabcolsep}{1.2pt}
  \begin{tabular*}{\textwidth}{@{\extracolsep{\fill}}p{0.25\textwidth}cccccccc@{}}
    \toprule
    & \multicolumn{3}{c}{\textbf{Tracker fall-only SR (\%) $\uparrow$}} &
    \multicolumn{5}{c}{\textbf{Generator}} \\
    \cmidrule(lr){2-4}\cmidrule(lr){5-9}
    \textbf{Variant} & \textbf{LAFAN1} & \textbf{AMASS-test} &
    \textbf{Wild-G1} &
    \metrichead{\textbf{IsaacLab}\\\textbf{Succ. $\uparrow$}} &
    \metrichead{\textbf{TMR}\\\textbf{R@1 $\uparrow$}} &
    \metrichead{\textbf{$E_{\rm joint}$}\\\textbf{$\downarrow$}} &
    \metrichead{\textbf{$E_{\rm key}$}\\\textbf{$\downarrow$}} &
    \metrichead{\textbf{TMR}\\\textbf{FID $\downarrow$}} \\
    \midrule
    \multicolumn{9}{l}{\textbf{\textit{(a) Update directions}}} \\
    No update ($G_0,T_0$) & 75.0 & \textbf{81.2} & 45.9 &
    92.58 & 0.774 & \textbf{0.159} & 0.410 & 0.023 \\
    Tracker only & \textbf{77.5} & 79.0 & 46.8 &
    92.58 & 0.774 & \textbf{0.159} & 0.410 & 0.023 \\
    Generator only & 75.0 & \textbf{81.2} & 45.9 &
    \textbf{93.95} & \textbf{0.783} & \textbf{0.157} &
    \textbf{0.381} & \textbf{0.020} \\
    \textbf{\ours{}} &
    75.0 & \textbf{81.2} & \textbf{47.3} & 93.55 & 0.782 & 0.160 &
    0.399 & \textbf{0.020} \\
    \midrule
    \multicolumn{9}{l}{\textbf{\textit{(b) Feedback and objective}}} \\
    w/o exec. reward &
    \textbf{77.5} & 78.9 & 46.1 & 92.71 & 0.778 & 0.161 & 0.408 & 0.024 \\
    Success only &
    76.3 & 80.2 & 46.7 & 93.12 & \textbf{0.780} & 0.160 &
    0.406 & \textbf{0.021} \\
    Current judge &
    74.8 & 79.6 & 46.4 & 93.01 & 0.777 & 0.163 & 0.412 & 0.026 \\
    RW-SFT &
    76.1 & 79.8 & 46.6 & \textbf{94.10} & 0.769 & \textbf{0.158} &
    \textbf{0.389} & 0.029 \\
    DPO &
    75.2 & 80.5 & 46.9 & 93.36 & 0.776 & 0.159 &
    0.402 & 0.023 \\
    \textbf{\ours{}} &
    75.0 & \textbf{81.2} & \textbf{47.3} & 93.55 & \textbf{0.782} &
    0.160 & 0.399 & \textbf{0.020} \\
    \bottomrule
  \end{tabular*}}
  \caption{Matched closed-loop ablations from the ProtoMotions initialization.
  Panel (a) separates generator and tracker update directions; panel (b)
  compares the full configuration with variants that change only the execution
  feedback, judge timing, or generator objective.  Tracker only freezes $G_0$,
  Generator only freezes $T_0$, and \ours{} uses FlowGRPO with a lagged judge.
  Current judge uses the adapting trainee, and RW-SFT denotes reward-weighted
  SFT.
  All variants share prompts, candidate groups, accepted-motion and
  tracker-transition budgets, validation, and frozen evaluators.  Generator
  metrics use the fixed held-out prompt suite. Bold marks the best value within
  each panel, including ties.}
  \label{tab:ablation-replay}
\end{table*}

\begin{table*}[!t]
  \centering
  {\small
  \setlength{\tabcolsep}{2.2pt}
  \begin{tabular*}{\textwidth}{@{\extracolsep{\fill}}p{0.30\textwidth}ccccccc@{}}
    \toprule
    \textbf{Variant} & \textbf{$\lambda_{\rm KL}$} &
    \textbf{$\lambda_{\rm GT}$} &
    \metrichead{\textbf{IsaacLab}\\\textbf{Succ. $\uparrow$}} &
    \metrichead{\textbf{$E_{\rm key}$}\\\textbf{$\downarrow$}} &
    \metrichead{\textbf{TMR}\\\textbf{R@1 $\uparrow$}} &
    \metrichead{\textbf{TMR}\\\textbf{FID $\downarrow$}} &
    \textbf{Diversity} \\
    \midrule
    w/o KL & 0 & 1.0 & 93.84 & 0.389 & 0.773 & 0.028 & 36.102 \\
    w/o GT & 0.02 & 0 & 94.07 & 0.382 & 0.768 &
    0.034 & 35.812 \\
    w/o both & 0 & 0 & \textbf{94.62} &
    \textbf{0.369} & 0.749 & 0.051 & 34.684 \\
    \textbf{\ours{}} & 0.02 & 1.0 & 93.55 & 0.399 & \textbf{0.782} &
    \textbf{0.020} & \textbf{36.373} \\
    \bottomrule
  \end{tabular*}}
  \caption{Matched generator-retention ablations for the ProtoMotions branch.
  $\lambda_{\rm KL}$ weights the frozen-generator penalty and
  $\lambda_{\rm GT}$ weights the supervised flow-matching update applied every
  two FlowGRPO iterations in the full configuration.  The controls remove
  either retention component or both while retaining the prompts, online-loop
  schedule, optimization budget, and frozen evaluators. Bold marks the strongest
  physical value or the best-preserved semantic/distribution value.}
  \label{tab:ablation-retention}
\end{table*}

\subsection{Control Experiments}
\label{app:control-experiments}

The controls in the main tables cover reference-only continuation, tracker-only
frozen-$G_0$ replay, tracker-filtered SFT, and frozen-SONIC FlowGRPO.  Together,
they provide direct comparisons for additional tracker optimization, static
replay, and generator feedback.  Table~\ref{tab:ablation-replay} completes the
matched ProtoMotions comparison by separating update directions in panel (a)
and varying the execution score, reward judge, and generator objective in
panel (b).

\paragraph{Generator-side effect.}
Relative to $G_0$ in the main-paper generator table, generator-only
feedback raises frozen-SONIC execution success from 92.58 to
93.95 and reduces $E_\mathrm{key}$ from 0.410 to 0.381.  At the same time,
$E_\mathrm{joint}$ decreases from 0.159 to 0.157, while TMR R@1 and FID
change from 0.774/0.023 to 0.783/0.020.  The fixed strong-tracker control improves
$E_\mathrm{key}$ to 0.363 but lowers success and R@1 to 90.92/0.767 and raises
FID to 0.027, whereas the online SONIC branch reaches 94.43 success,
$E_\mathrm{key}=0.325$, R@1 $=0.783$, and FID $=0.020$.  These trends support
the combined design of execution-grounded generator updates, evolving tracker
feedback, and KL/rehearsal anchors that preserve the pretrained motion
distribution.

\paragraph{Tracker-side effect.}
The tracker controls explain why generated supervision is updated online rather
than stored as a fixed replay bank.  Frozen-$G_0$ replay provides useful but
split-dependent gains.  With SONIC, the full loop improves its three success
rates from 85.0/78.3/45.5 to 90.0/79.7/48.0 relative to frozen-$G_0$ replay,
while reducing MPJPE from 133.3 to 124.1~mm and $E_g$ from 847.6 to
807.2~mm.  With ProtoMotions, it improves AMASS-test and Wild-G1 success and
reduces MPJPE.  Reference-only continuation likewise does not reproduce this
cross-split pattern.  These controls support the generated-reference curriculum
and the central choice to evolve generator feedback and tracker supervision
online rather than relying only on additional tracker updates or static replay.

\paragraph{Execution-feedback and objective controls.}
Panel (b) of Table~\ref{tab:ablation-replay} maps each comparison directly to a
method choice.  Removing execution reward leaves the generator at 92.71
success, 0.778 R@1, and 0.024 FID, while AMASS-test/Wild-G1 SR reach
78.9/46.1.  Success-only feedback improves these values to
93.12/0.780/0.021 and 80.2/46.7, respectively, but remains below the dense-score
configuration.  Using the current trainee as sole judge lowers cross-split SR
to 74.8/79.6/46.4 and degrades $E_\mathrm{joint}/E_\mathrm{key}$ to
0.163/0.412.  Reward-weighted SFT obtains the highest raw success (94.10) and
lowest $E_\mathrm{joint}/E_\mathrm{key}$ (0.158/0.389), but reduces R@1 to
0.769 and raises FID to 0.029.  Diffusion-DPO provides a more balanced control
(93.36 success, 0.776 R@1, 0.023 FID) and reaches 80.5/46.9 on
AMASS-test/Wild-G1, but remains below FlowGRPO on execution success, key-body
alignment, R@1/FID, and the two target-split coverages.  Because these controls
retain the same online samples, rewards, and update budget, the comparison
supports dense lagged feedback and FlowGRPO as complementary choices rather
than consequences of extra data or optimization.

\paragraph{Generator-retention controls.}
Table~\ref{tab:ablation-retention} connects the preservation mechanisms to the
generator-side claim.  Removing either anchor improves raw executability but
reduces TMR retrieval, increases FID, and lowers diversity; the degradation is
largest when both anchors are removed.  The full configuration therefore
retains most of the physical benefit while providing the strongest
physical--semantic balance.

\subsection{Qualitative and Distribution-Gap Analysis}
\label{app:qualitative-analysis}

\paragraph{Paired qualitative evidence.}
The paired visualizations expose the complementary effects targeted by our
alternating updates.  Under the same prompt, length, and sampling noise,
replacing $G_0$ with the \ours{} generator turns a reference that the same frozen
SONIC policy fails to execute into one that it completes.  Conversely, for the
shown held-out Wild-G1 reference, both the initial tracker and the
frozen-$G_0$ replay control fail, whereas the online \ours{} tracker completes
the motion.  Together with the main-paper generator and tracker tables, these
cases visualize the role of each branch:
tracker rollouts guide the generator toward robot-compatible references, while
an evolving generated-reference curriculum extends tracker coverage beyond
static replay.  All snapshots render logged poses on the Unitree G1 STL model.

\paragraph{Reference--execution diagnostics.}
The main-paper reference--execution figure analyzes a diagnostic high-fidelity
subset selected by predefined execution-fidelity thresholds.  Shared PCA of
9,332 paired windows exhibits broad reference--execution overlap, with its first
two components explaining 61.1\% of pooled variance.  The shared structure
supports robot-native initialization and rehearsal, but a systematic local gap
remains: a motion-case-grouped classifier in the full descriptor space reaches
an AUC of $0.996\pm0.004$, compared with $0.505\pm0.006$ under shuffled labels.
Across all 10,369 diagnostic windows, descriptor densities localize the largest
shifts to body jerk, foot skating, root height, and foot clearance, with a
smaller change in peak root speed.  These results support the central method
choice to evaluate motions through closed-loop execution and use a dense
trajectory-level score, rather than treating robot-space references or a fixed
generated pool as executable by construction.

\section{Limitations}
\label{app:limitations}

\ours{} depends on retargeting, simulation, and tracker quality.  We therefore
evaluate held-out GMD-retargeted G1 references, robot-native G1 motions produced
by the generator, and their tracker executions separately.  The evaluation is
scoped to Unitree G1 in simulation.  Because physical feedback can favor
conservative motions, we report text alignment and diversity alongside physical
metrics.

\end{document}